\documentclass[conference]{IEEEtran}
\IEEEoverridecommandlockouts
\usepackage{cite}
\usepackage{amsmath,amssymb,amsfonts}
\usepackage{algorithmic}
\usepackage{graphicx}
\usepackage{textcomp}

\usepackage{xcolor}
\def\BibTeX{{\rm B\kern-.05em{\sc i\kern-.025em b}\kern-.08em
    T\kern-.1667em\lower.7ex\hbox{E}\kern-.125emX}}
\usepackage{comment}
\usepackage{mathtools}
\usepackage{adjustbox}
\usepackage{booktabs}
\usepackage{multirow}
\usepackage{bbm}
\usepackage{makecell}

\newcommand{\stdsize}{\fontsize{6}{6}\selectfont}
\newcommand{\num}[2]{#1 \stdsize{$\pm$ #2}}
\newcommand{\bum}[2]{\textbf{#1} \stdsize{$\pm$ #2}}
\newcommand{\uum}[2]{\underline{#1} \stdsize{$\pm$ #2}}

\newcommand{\alg}[1]{\textsc{#1}}
\DeclareMathOperator*{\argmax}{arg\,max}
\DeclareMathOperator*{\argmin}{arg\,min}
\newcommand{\N}{\mathbb{N}}
\newcommand\norm[1]{\left\lVert#1\right\rVert}

\def\onedot{.}
\def\eg{\emph{e.g}\onedot} 
\def\ie{\emph{i.e}\onedot} 

\usepackage[capitalize]{cleveref}
\crefname{section}{Sec.}{Secs.}
\Crefname{section}{Section}{Sections}
\Crefname{table}{Table}{Tables}
\crefname{table}{Tab.}{Tabs.}

\begin{document}

\title{FedDiverse: Tackling Data Heterogeneity in Federated Learning with Diversity-Driven Client Selection\\
\thanks{This research was funded by the European Union.  Views and opinions expressed are however those of the author(s) only and do not necessarily reflect those of the European Union or the European Health and Digital Executive Agency (HaDEA). Neither the European Union nor the granting authority can be held responsible for them. G.D.N. and N.O. have been partially supported by funding received at the ELLIS Unit Alicante Foundation by the European Commission under the Horizon Europe Programme - Grant Agreement 101120237 - ELIAS, and a nominal grant from the Regional Government of Valencia in Spain (Convenio Singular signed with Generalitat Valenciana, Conselleria de Innovación, Industria, Comercio y Turismo, Dirección General de Innovación). G.D.N. is also funded by a grant by the Banco Sabadell Foundation. G.D.N., E.F., Y.J.N., and N.Q. have been supported in part by the European Research Council under the European Union’s Horizon 2020 research and innovation programme Grant Agreement no. 851538 - BayesianGDPR. N.Q. has been supported in part by the Horizon Europe research and innovation programme Grant Agreement no. 101120763 - TANGO. N.Q. is also supported by BCAM Severo Ochoa accreditation CEX2021-001142-S/MICIN/AEI/10.13039/501100011033.}
}

\author{
\IEEEauthorblockN{Gergely D. Németh}
\IEEEauthorblockA{
\textit{ELLIS Alicante}\\
Alicante, Spain \\
neged.ng@gmail.com}
\and
\IEEEauthorblockN{Eros Fanì}
\IEEEauthorblockA{\textit{Polytechnic University of Turin,} \\
Turin, Italy \\
eros.fani@polito.it}
\and
\IEEEauthorblockN{Yeat Jeng Ng}
\IEEEauthorblockA{
\textit{University of Sussex}\\
Brighton, UK \\
Y.Ng@sussex.ac.uk}
\and
\IEEEauthorblockN{Barbara Caputo}
\IEEEauthorblockA{\textit{Polytechnic University of Turin,} \\
Turin, Italy \\
barbara.caputo@polito.it}
\and
\IEEEauthorblockN{Miguel Ángel Lozano}
\IEEEauthorblockA{
\textit{University of Alicante}\\
Alicante, Spain \\
malozano@gcloud.ua.es}
\and
\IEEEauthorblockN{Nuria Oliver}
\IEEEauthorblockA{
\textit{ELLIS Alicante}\\
Alicante, Spain \\
nuria@ellisalicante.org}
\and
\IEEEauthorblockN{Novi Quadrianto}
\IEEEauthorblockA{\textit{University of Sussex}, \\
Brighton, UK \\
N.Quadrianto@sussex.ac.uk}
}

\maketitle

\begin{abstract}
    Federated Learning (FL) enables decentralized training of machine learning models on distributed data while preserving privacy. However, in real-world FL settings, client data is often non-identically distributed and imbalanced, resulting in statistical data heterogeneity which impacts the generalization capabilities of the server's model across clients, slows convergence and reduces performance. In this paper, we address this challenge by proposing first a characterization of statistical data heterogeneity by means of 6 metrics of global and client attribute imbalance, class imbalance, and spurious correlations. Next, we create and share 7 computer vision datasets for binary and multiclass image classification tasks in Federated Learning that cover a broad range of statistical data heterogeneity and hence simulate real-world situations. Finally, we 
    propose \alg{FedDiverse}, a novel client selection algorithm in FL which is designed to manage and leverage data heterogeneity across clients by promoting collaboration between clients with complementary data distributions. 
    Experiments on the seven proposed FL datasets demonstrate \alg{FedDiverse}'s effectiveness in enhancing the performance and robustness of a variety of FL methods while having low communication and computational overhead. 
\end{abstract}

\begin{IEEEkeywords}
component, formatting, style, styling, insert
\end{IEEEkeywords}

\begin{figure}
    \centering
    \includegraphics[width=0.8\linewidth]{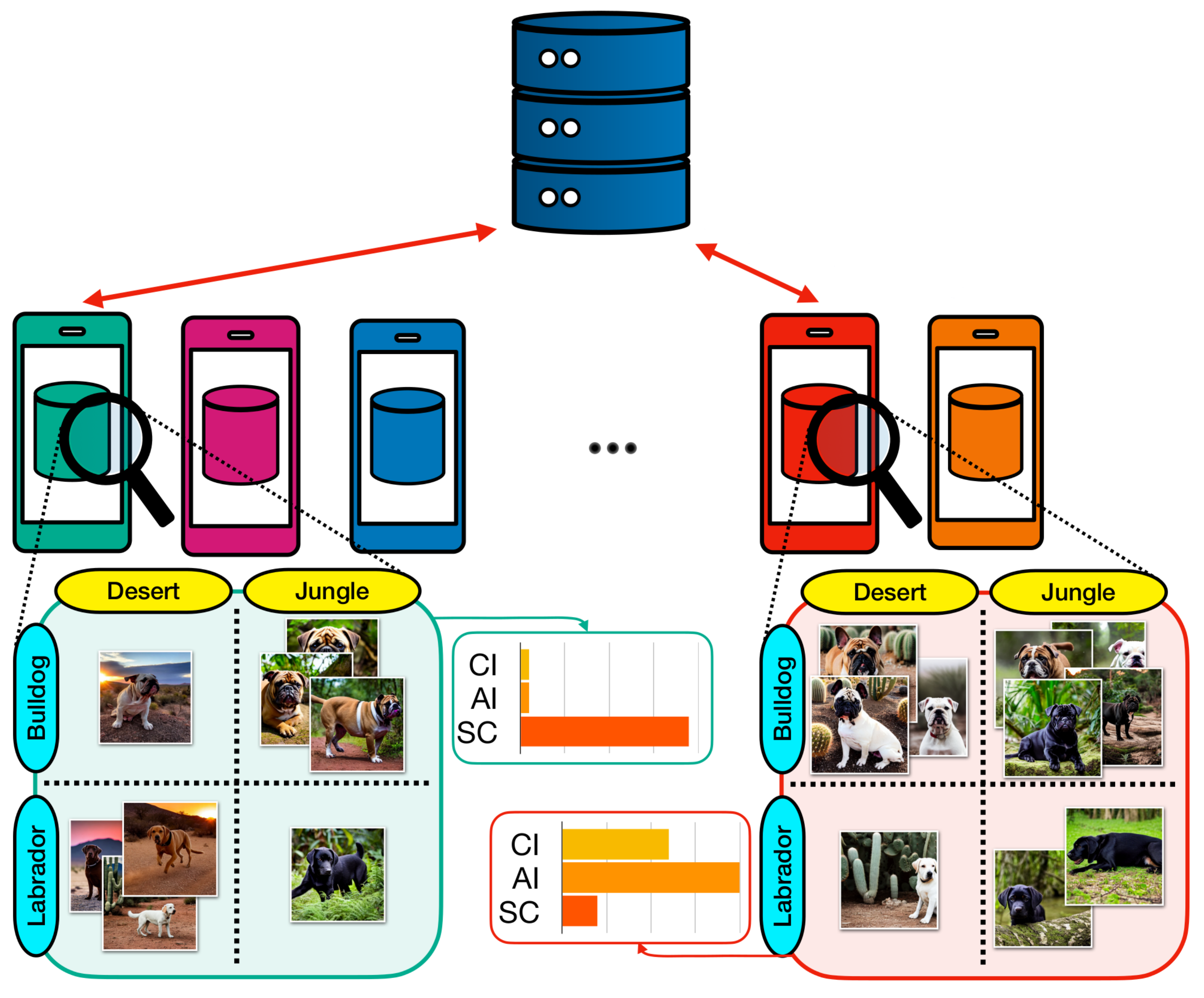}
    \caption{Visual representation of \textsc{FedDiverse}. CI, AI, and SC stand for Class Imbalance, Attribute Imbalance, and Spurious Correlation in the clients' data distributions. Observe the statistical data heterogeneity in the selected clients (turquoise and red). \alg{FedDiverse} automatically selects clients with a diversity of local statistics to learn a global model that is resilient to statistical data heterogeneity.}
    \label{fig:teaser}
\end{figure}

\section{Introduction}
\label{sec:intro}


In centralized machine learning, all training data is shared with a central server, posing privacy, regulatory, and ethical concerns, especially for sensitive data \cite{li2020federated}. Federated learning (FL) \cite{mcmahan2017communication} aims to address these concerns by enabling decentralized, privacy-preserving model training without transferring raw data. In FL, models are trained collaboratively across distributed \textit{clients}. In each training round: (1) the server shares the global model parameters with selected clients; (2) the clients perform local training using those parameters; and (3) the clients send the updated model parameters back to the server for aggregation. This decentralized process maintains data privacy while improving the global model.

In real-world FL scenarios, client data is often shaped by local factors such as differing user behaviors \cite{tan2022towards}, context-specific data collection environments \cite{fallah2020personalized}, and socio-economic or cultural biases \cite{brisimi2018federated}, resulting in  \emph{statistical data heterogeneity}, where data across different clients is non-independent and identically distributed (non-IID) and imbalanced. Statistical data heterogeneity hampers the generalization capabilities of the server's model across clients, slowing convergence and reducing performance \cite{li2019convergence}.

Previous studies in FL have addressed statistical data heterogeneity from an algorithmic perspective, providing convergence theorems, analyzing computational costs, and proposing solutions to mitigate its effects  \cite{li2019convergence, li2020federated}. However, there is a lack of fine-grained analyses of this problem. In this paper, we address this by decomposing 
the \emph{attribute}-\emph{target label} relationships to identify three types of data heterogeneity: (1) \emph{class imbalance} (CI), when target labels have asymmetric distributions; (2) \emph{attribute imbalance} (AI), when attributes exhibit imbalanced distributions; and (3) \emph{spurious correlations} (SC), which emerge when the model learns misleading correlations between a non-discriminative attribute, such as the background, and the target label. These three types of data heterogeneity pose a challenge in both centralized \cite{yurtsever2020survey,yang2023change} and federated \cite{kairouz2021advances,mora2024enhancing} learning. 

Previous work in centralized machine learning has shown that CI, AI, and SC often arise when data is limited or lacks sufficient diversity \cite{ye2024spurious, geirhos2020shortcut}. Thus, a typical solution consists of using an additional and diverse yet unlabeled dataset --called ``validation'', ``target'' or ``deployment'' dataset-- to perform self-training (\eg \cite{liu2021just,CheWeiKumMa20}) or to learn a representation that is invariant to attributes (\eg \cite{TraCreKilLocetal21}).

In FL, the \emph{diversity} of the client data could be leveraged to devise client selection methods that mitigate the effects of CI, AI, and SC. By prioritizing clients with complementary data distributions, the server's model is exposed to diverse training patterns without accessing raw data, enhancing generalization while preserving privacy. In this paper, we develop this idea and address the challenge of statistical data heterogeneity in FL by proposing a novel client selection algorithm called \alg{FedDiverse} that leverages of diversity in client data distributions. We empirically evaluate \alg{FedDiverse} on 7 computer vision datasets that exhibit varying levels of CI, AI and SC, leading to the following \textbf{contributions}:

(1) We propose a fine-grained analysis of statistical data heterogeneity in FL by means of 6 metrics; 

(2) We introduce and share 7 FL datasets for binary and multiclass image classification tasks that cover a broad range statistical data heterogeneity; 

(3) We present and evaluate \alg{FedDiverse}, illustrated in \cref{fig:teaser}, a novel client selection method that is designed to mitigate the impact of statistical data heterogeneity (CI, AI and SC) in FL training while ensuring the privacy of clients and respecting the resource-constrained nature of each client. 

\section{Related Work}
\label{sec:related}

\subsection{Data Heterogeneity in Federated Learning}

Statistical heterogeneity or non-IID data is a major concern in FL because it can hinder the training process, leading to poor generalization and slow and unstable convergence \cite{kairouz2021advances}. Various methods have been proposed to address this issue \cite{mora2024enhancing}. Some approaches add regularization terms to align local updates with the global model, such as 
\alg{FedDyn} \cite{acar2021federated} 
and \alg{FedProx} \cite{li2020federated}, 
while other methods aim to reduce variance between client updates, such as 
\alg{MOON} \cite{li2021model}, 
and \alg{FedFM}~\cite{ye2023fedfm}. 
In other approaches, the clients share additional information with the server that reveals information about their statistical data heterogeneity. In \alg{pow-d}~\cite{cho2022towards}, clients share the average loss of the previous global model applied to their local data; in \alg{IGPE}~\cite{zhang2024addressing} they share averaged network embeddings;  and in \alg{FedAF}~\cite{wang2024aggregation} they share synthetic data. 
Finally, optimization-based server-side methods, such as \alg{FedAvgM}~\cite{hsu2019measuring}, 
and \alg{FedOpt} \cite{reddi2020adaptive}, employ adaptive learning rates at the server to manage statistical diversity among clients.

However, none of these strategies explicitly addresses the challenge posed by spurious correlations in client data, leaving room for improvement.

\subsection{Spurious Correlations in Centralized ML}

Spurious correlations can significantly hinder robustness and generalization in machine learning \cite{ye2024spurious, geirhos2020shortcut}. Proposed solutions to this problem fall into two main categories. The first category \cite{SKHL2020,YWL+2022} unrealistically assumes that spurious attributes are known or partially labeled, enabling models to reduce reliance on these attributes by re-weighting samples or modifying training processes. These methods often require that data groups or environments be explicitly defined to minimize spurious dependencies. 

The second category does not assume prior knowledge of spurious attributes. Instead, models are designed to automatically distinguish meaningful patterns from spurious ones, often using techniques such as adversarial training \cite{kim2019learning, clark-etal-2019-dont} or counterfactual data augmentation \cite{wang2019balanced}.
For example, \alg{LfF} trains two models concurrently: a biased model to capture dataset biases and a debiased one trained on re-weighted samples influenced by the biased model's predictions \cite{nam2020learning}; and \alg{Just-Train-Twice} initially identifies ``failure" cases where the model misclassifies, then increases the weights of these cases in a second training phase to improve robustness against spurious features \cite{liu2021just}. 

Even though spurious correlations have been sparsely studied in the FL literature, recent research has begun to explore this challenge. To the best of our knowledge,  \cite{wangpersonalized} is the first piece of work to tackle spurious correlations in FL by investigating personalization such that models are tailored to the individual clients' data. 
In contrast, we aim to learn a single global model that remains robust to spurious correlations across all client distributions, achieving strong generalization performance for all clients. 

\subsection{Client Selection and Weighting in FL}

Client selection and client weighting are two primary strategies in FL to manage client contributions during training and mitigate the challenges posed by heterogeneous data~\cite{nemeth2022snapshot}. In client selection, which is especially relevant in resource-constrained settings, only a subset of clients participate in each training round to reduce communication and resource demands, improving training efficiency. Conversely, client weighting includes all clients in each round but adjusts their influence on the global model by means of a weight, aiming to accelerate convergence and performance~\cite{cho2022towards}. Both strategies support fairness~\cite{ caldas2018expanding} and security~\cite{rodriguez2022dynamic}, mitigating effects from clients with unreliable or adversarial data. 

Client selection or client weighting strategies address the challenge of statistical heterogeneity in FL by prioritizing or scaling the client contributions based on data quality and relevance. In the client selection category, methods like \alg{FedPNS}~\cite{wu2022node} 
and \alg{pow-d}~\cite{cho2022towards} prioritize clients that are expected to contribute significantly to model accuracy, either through gradient similarity to the average model gradient or by selecting clients whose data produces high loss on the server's model. \alg{Fed-CBS} aims to reduce the class-imbalance by selecting the clients that will generate a more class-balanced grouped dataset \cite{zhang2023fed}.

Client clustering is a common technique for selecting clients that represent groups that share similar data distributions%
\footnote{We exclude works referred as clustered federated learning (FL), where each client cluster trains a separate model personalized to the data distribution of that cluster, as our aim is to train one robust model shared by all clients.}.
Server-side clustering methods typically consider the similarity of the client gradient updates as a proxy of the similarity between their data distributions (e.g., \alg{FCCPS}~\cite{xin2022federated}) 
or their projection into a lower dimension for compression (e.g. \alg{HCSFed}~\cite{song2023fast}). In addition, clients can send metrics that describe the statistical heterogeneity of their local data, such as entropy in \alg{HiCS-FL}~\cite{chen2025heterogeneity}. Sharing the full characteristics of the client data distribution with the server has also been investigated~\cite{wolfrath2022haccs}, yet it could be considered a privacy violation~\cite{chen2025heterogeneity}, and it is typically unknown for spurious correlations.
Finally, client weighting methods, such as \alg{CI-MR}~\cite{song2019profit},  \alg{FMore}~\cite{zeng2020fmore} and \alg{FedNova}~\cite{wang2020tackling}, reward clients with high-value data or normalize updates to counter statistical heterogeneity. 

Although most existing methods address non-IID data in FL through class imbalance, we study other types of statistical heterogeneity, such as attribute imbalance and spurious correlations, as described next.

\section{A Framework of Data Heterogeneity in FL}
\label{sec:background}


\subsection{Background and Problem Setup}
\label{sec:problemsetup}

Let $f: \mathcal X \to \mathcal Y$ be a predictor function parameterized by $\theta \in \Theta$, where $\mathcal X$ is the feature space, $\mathcal Y$ is the output space, and $\Theta$ is the parameter space.
We assume that the feature space consists of two subspaces: $\mathcal X \subseteq \mathcal X_{y} \times \mathcal X_{a}$, where $\mathcal X_{y}$ and $\mathcal X_{a}$ represent the \textit{task-intrinsic} and the \textit{attribute} feature spaces, respectively.
The class label $y \in \mathcal Y$ of a sample $x \coloneqq (x_{y}, x_{a})$ is determined by the discriminative feature $x_{y}$ whereas the attribute label $a \in \mathcal A$ is determined by the attribute feature $x_{a}$, where $\mathcal{A}$ is the space of attributes.
The training dataset $D$ consists of $n$ feature-target sample pairs\footnote{In this work, we assume that the labeling of the attribute is not available in the training set.}, $D = \left\{ (x_i,y_i) \right\}_{i=1}^n$, where each sample is identically and independently drawn from the training distribution $\mathbb P_{\text{tr}}$.

In a FL scenario, the dataset $D$ composed of $n$ samples is split across $K$ clients $k \in \mathcal{K}$. In other words, each client $k$ has access to a local, private dataset $D_k$ such that $D = \bigcup_{k \in \mathcal{K}} D_k$, $\left | D_k \right | \coloneqq n_k$, and $\sum_{k \in \mathcal{K}} n_k = n$, which cannot be accessed neither by the server $\mathcal{S}$ nor by any other client $j \neq k \in \mathcal{K}$. The FL objective is to find optimal parameters $\theta^* \in \Theta$ by solving the following problem:

\begin{equation}
    \label{eq:fl_objective}
    \theta^* = \argmin_{\theta \in \Theta} \sum_{k \in \mathcal{K}} \frac{n_k}{n} \mathcal{L}_k(\theta),
\end{equation}

\noindent where $\mathcal{L}_k(\theta) = \sum_{(x, y) \in D_k} \ell(f(x; \theta), y)$ and $\ell$ is any loss function.

In practice, federated learning is orchestrated by a central server $\mathcal{S}$, which schedules the training into $T > 0$ rounds. During each round $t$, $0 < t \leq T$, a set $\mathcal{K}' \subseteq \mathcal{K}$ of clients is sampled by $\mathcal{S}$ and shares the current global parameters $\theta^t$ with them. Then, each client $k \in \mathcal{K}'$ initializes its local model with the received parameters and trains it using its local dataset $D_k$, obtaining new parameters $\theta^{t+1}_k$. Finally, each sampled client shares its parameters with the server $\mathcal{S}$, where they are aggregated to form new global parameters $\theta^{t+1}$. In the case of the standard \alg{FedAvg} \cite{mcmahan2017communication}, this parameter aggregation is performed by computing the weighted mean: 
$\theta^{t+1} = \sum_{k \in \mathcal{K}'} \frac{n_k}{n} \theta_k^{t+1}$. This procedure is repeated for several rounds until convergence.

\subsection{Statistical Data Heterogeneity}
\label{sec:subpop}

Statistical data heterogeneity emerges when there is a subpopulation shift, \ie, when the representation of subpopulations differs between the training $\mathbb P_{tr}$ and the test $\mathbb P_{te}$ distributions.
Here, subpopulations are defined by the target labels and the attributes, $\mathcal Y \times \mathcal A$.
We consider three types of statistical data heterogeneity: 

\textbf{Class Imbalance (CI):}
The distribution of the target labels $y$ is different between the training and test distributions, such that certain classes are more prevalent in the training than in the test sets, \ie:
 $   \mathbb P_{\text{tr}}(Y = y) \gg \mathbb P_{\text{tr}}(Y = y')$
for some $y,y' \in \mathcal Y$ where $y \neq y'$. CI can yield a biased classifier that performs poorly in samples from the minority class.

\textbf{Attribute Imbalance (AI):} The probability of occurrence of a certain attribute $a'$ in the training set is much smaller than other attributes $a \in \mathcal A$ and this disparity in prevalence does not hold in the test distribution, \ie,
    $\mathbb P_{\text{tr}}(A = a) \gg \mathbb P_{\text{tr}}(A = a')$.
AI can produce a biased classifier towards the majority attribute $a$.

\textbf{Spurious Correlation (SC):}
There is a statistical dependency between the class $Y$ and the attribute $A$ in the training distribution, which does not exist in the test distribution, \ie, $
    \mathbb P_{\text{tr}}(Y=y \mid A=a) \gg \mathbb P_{\text{tr}}(Y=y) \gg \mathbb P_{\text{tr}}(Y=y \mid A=a')$, 
for some $y \in \mathcal Y$ and $a, a' \in \mathcal A$.
This spurious dependency can cause a classifier to perform well on samples where the spurious relationship holds (\eg, $(Y=y, A=a)$), but to underperform where the relationship does not hold (\eg, $(Y=y, A=a')$).


\subsection{Data Heterogeneity Metrics}
\label{sec:flmetrics}

\textbf{Centralized Metrics.}
To measure the degree of statistical data heterogeneity in dataset $D$, we adopt the metrics proposed in \cite{yang2023change}:
\begin{align}
    \Delta_{\text{CI}}(D) & = 1- H(Y) / \log |\mathcal Y| \label{eq:ci} \\
    \Delta_{\text{AI}}(D) & = 1 -H(A) / \log |\mathcal A| \label{eq:ai} \\
    \Delta_{\text{SC}}(D) & = 2 I(Y;A) / (H(Y) + H(A)) \label{eq:sc}
\end{align}

\noindent where $H$ and $I$ are the entropy and mutual information with respect to the empirical distribution of the dataset, respectively. Each metric is bounded within $[0,1]$. 

\textbf{Federated Learning Metrics.}
We present six metrics --three global and three local-- that characterize statistical data heterogeneity in FL, expanding the previously presented metrics for centralized learning. 

\textit{Global FL metrics.} In FL, when the metrics in \Cref{eq:ci,eq:ai,eq:sc} are computed on the union of the clients' datasets, \ie $D = \bigcup_{k \in \mathcal{K}} D_k$, they provide a global understanding of the severity of CI, AI and SC, namely: 
\begin{align}
    &\resizebox{.73\linewidth}{!}{$
    \text{\textbf{Global Class Imbalance: }} GCI = \Delta_{\text{CI}}(D)$} \label{eq:gci} \\
    &\resizebox{.80\linewidth}{!}{$
    \text{\textbf{Global Attribute Imbalance: }} GAI = \Delta_{\text{AI}}(D)$} \label{eq:gai} \\
    &\resizebox{.82\linewidth}{!}{$
    \text{\textbf{Global Spurious Correlation: }} GSC = \Delta_{\text{SC}}(D)$} \label{eq:gsc}
\end{align}

\textit{Client FL metrics.} The global FL metrics fail to capture the heterogeneity present in the datasets of individual clients. To this end, we propose three additional client metrics, where the local values of CI, AI and SC are averaged across all the $K$ clients:
\begin{align}
    &\resizebox{.73\linewidth}{!}{$
    \text{\textbf{Client Class Imbalance: }} CCI = \frac{1}{K} \sum_{k \in \mathcal{K}} \Delta_{\text{CI}}(D_k)
    $} \label{eq:lci} \\
    &\resizebox{.80\linewidth}{!}{$
    \text{\textbf{Client Attribute Imbalance: }} CAI = \frac{1}{K} \sum_{k \in \mathcal{K}} \Delta_{\text{AI}}(D_k)
    $} \label{eq:lai} \\
    &\resizebox{.82\linewidth}{!}{$
    \text{\textbf{Client Spurious Correlation: }} CSC = \frac{1}{K} \sum_{k \in \mathcal{K}} \Delta_{\text{SC}}(D_k)
    $} \label{eq:lsc}
\end{align}

In practice, data heterogeneity 
often consists of a mixture of CI, AI and SC in the data distributions of different clients, as shown in \cref{fig:teaser}
where Bulldog/Labrador is the target label and Desert/Jungle as the non-discriminative attribute in the image classification task.








\section{Client Selection via \alg{FedDiverse}}
\label{sec:method}

\begin{figure*}[tp]
    \centering
    \includegraphics[width=0.8\linewidth]{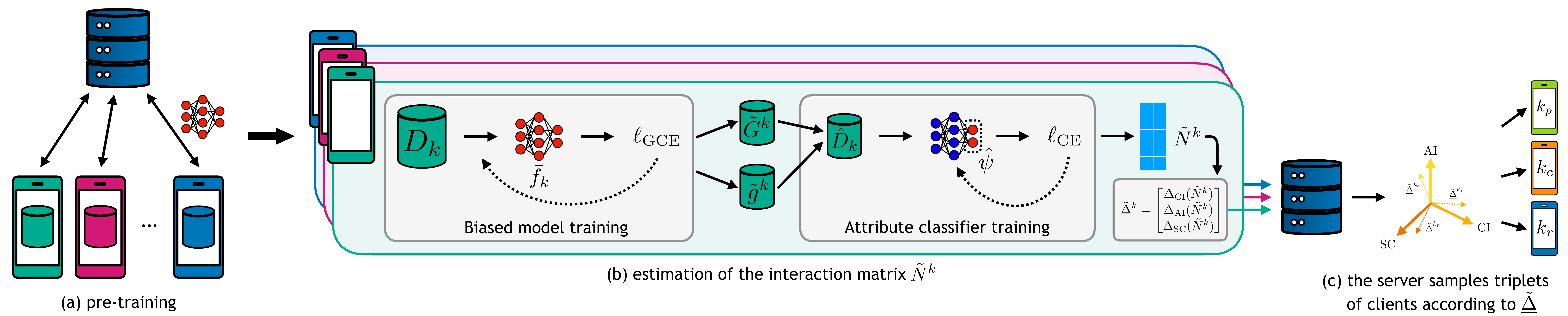}
    \caption{Main steps of \alg{FedDiverse}. First (a), there is a phase of standard federated model pre-training. Second (b), the clients estimate their interaction matrices and, from them, their data heterogeneity triplets, which they share with the server. Finally, (c), the server uses the received triplets to perform client selection. Learnable parameters are marked in red, while fixed parameters are in blue.}
    \label{fig:method}
\end{figure*}
 
The proposed \alg{FedDiverse} method consists of two components, illustrated in \cref{fig:method} and described next. First, an approach to estimate the statistical data heterogeneity in the clients, characterized by their local CI, AI and SC (\cref{sec:intmatrix}). Second, a client selection strategy designed to include diverse clients in each round from the perspective of their statistical data heterogeneity (\cref{sec:feddiverse}). 

\subsection{Estimation of the Statistical Data Heterogeneity}
\label{sec:intmatrix}

\paragraph{Preliminaries} The global \emph{interaction matrix} \( N \) represents the count of samples in a global dataset \( D \) by class \( \mathcal{Y} \) and attribute \( \mathcal{A} \). For each client \( k \), a local interaction matrix \( N^k \) captures its own non-normalized joint distribution of classes and attributes in their dataset $D_k$, such that \( N = \sum_{k \in \mathcal{K}} N^k \). Although clients cannot access the full distribution of their interaction matrices due to unknown attribute distributions, each can compute a marginal interaction vector $M^k \in \N^{|Y|} : M^k_y = \sum_{a \in \mathcal{A}} N^k_{ya}$, where $N^k_{ya}$ are the number of samples belonging to class $y \in Y$ and attribute $a \in A$ in the client's dataset $D_k$. Therefore, $M^k_y$ contains the distribution of the classes in dataset $D_k$.

The interaction matrix reflects the precise, non-normalized distributions of classes and attributes. In cases with strong spurious correlations, local models may rely on an attribute \( a \) which is the most correlated with the class \( y \) instead of intrinsic class features. For each client, the \emph{majority group} for a class \( y \), denoted as \( G^k_y \), includes the samples where the attribute \( a \) has the highest count in \( N^k \), and the \emph{minority group}, \( g^k_y \), includes samples where \( a \) has the lowest count. By aggregating these for each class, we define the majority group for client \( k \) as \( G^k \) and the minority group as \( g^k \). Because clients do not fully know the attribute function, they estimate these groups.

Finally, we assume that 
there are two attributes ($|\mathcal{A}| = 2$), the attribute set can be defined as \( A = \{ a_0, a_1 \} \). This structure allows clients to infer the minority group attributes for a class \( y \) once they know the majority group attribute.
Note that most datasets addressing SC or AI problems typically contain only two attributes (see Table 2 in \cite{yang2023change}).

\paragraph{Estimation of the interaction matrices}
Each client $k$ approximates \( N^k \) as \( \tilde{N}^k \) and uses this estimated matrix to compute its data heterogeneity triplet (DHT), $\Tilde{\Delta}^k = [\Delta_\text{CI}(\Tilde{N}^k), \Delta_\text{AI}(\Tilde{N}^k), \Delta_\text{SC}(\Tilde{N}^k)]^\top$\footnote{The metrics in \cref{eq:ci,eq:ai,eq:sc} can be equivalently calculated using the interaction matrix, as it fully describes the non-normalized joint distribution of classes and attributes.}. To preserve privacy, the clients only share the triplet with the server, which uses these triplets to select clients, as explained in Section \ref{sec:feddiverse}. 

In the following, we outline the three-step method adopted by the clients to estimate their interaction matrices \( \tilde{N}^k \) and hence their data heterogeneity triplets $\Tilde{\Delta}^k$. Note that this estimation is only performed once at the beginning of the FL training process.

\textbf{1.\,Pre-training:} A global pre-training phase is carried out for a small number of rounds $T_0$ using the \alg{FedAvg} algorithm, resulting in the global parameters $\theta^{T_0}$.

\textbf{2.\,Learning a Biased Model:} After pre-training, each client receives $\theta^{T_0}$ and overfits a local model called a \emph{biased model} $\bar{f}_k$ to its own data using the generalized cross-entropy loss function $\ell_{\text{GCE}}$ \cite{zhang2018generalized}. This loss function encourages the model to rely more heavily on easy-to-learn patterns, which are often associated with spurious correlations \cite{nam2020learning}. As a result, each client can distinguish between a majority group $G^k$ (where the majority of correctly predicted samples will belong) and a minority group $g^k$ (where the incorrectly predicted samples will mainly belong). The predicted majority and minority groups for class $y$ are denoted by $\Tilde{G}^k_y$ and $\Tilde{g}^k_y$, $\forall y \in \mathcal{Y}$, respectively. Given the nature of the $\ell_{GCE}$ loss, for $|\mathcal{Y}|>2$, we train one-vs-rest binary classifiers $\bar{f}_k^y$ for each $y\in\mathcal{Y}$ to determine $\Tilde{G}^k_y$ from the correctly predicted samples.

\textbf{3.\,Attribute classifier:} Using the biased model, clients label samples in the majority and minority groups, even though they lack information about the exact attribute labels. They identify a ``pivot class" which has the smallest difference in sample size between the predicted majority and minority groups, \emph{i.e.} $\Hat{y} = \argmin_{y \in \mathcal{Y}} \left | \left | \Tilde{G}^k_y \right | - \left | \Tilde{g}^k_y \right | \right |$. This class forms a new dataset 
$\Hat{D}_k$, which contains all the samples in $D_k$ whose class is $\Hat{y}$. Each client then trains an \emph{attribute classifier} $\Hat{\psi}$ locally on $\Hat{D}_k$ using cross-entropy loss to predict the attribute labels. This classifier yields an approximate interaction matrix $\Tilde{N}^k$ by predicting the attributes according to the attribute labels in $\Hat{D}_k$. Finally, each client computes their approximate DHT $\Tilde{\Delta}^k$ and sends it to the server $\mathcal{S}$. 

The server collects all the triplets sent by the clients in the \emph{approximate data heterogeneity matrix} $\Tilde{\Delta} \in [0, 1]^{3 \times K}$ where each column corresponds to one client $k$ and each row corresponds to the CI, AI, and SC components of the clients' $\Tilde{\Delta}^k$. 

Note that the final values of the scores in $\Tilde{\Delta}^k$ are the same independently of the specific labeling choice for the $\Hat{D}_k$ dataset, \ie,  clients could equivalently assign the attribute label 1 to the majority group samples and 0 to the minority group samples. Moreover, sharing $\Tilde{\Delta}^k$ does not disclose private information from the clients and only incurs negligible additional communication costs. Thus, this approach is suitable for resource-constrained scenarios. 

\subsection{Client Selection}
\label{sec:feddiverse}
The rationale of \alg{FedDiverse} is to sample clients with different types of statistical data heterogeneity (CI, AI and SC) in each round, leveraging it to achieve better generalization and robustness to real-word shifts \cite{zhang2023fed}.

\alg{FedDiverse}'s client selection is achieved by leveraging the information in the triplet $\Tilde{\Delta}^k$ received from each client and sampling clients to ensure diversity in the three dimensions of the triplets, \emph{i.e.}, selecting clients whose datasets exhibit a variety of CI, AI and SC. The client selection consists of the three steps described below.

\textbf{1.\,Probabilistic Selection (SC):} The first criterion for selecting a client is based on the presence of spurious correlations. The probability distribution $p_\text{SC}$ over all clients, based on the SC dimension of the data heterogeneity triplet (DHT) $\Tilde{\Delta}^k$ is given by:
    $p_\text{SC} = \frac{\Tilde{\Delta}_3}{\norm{\Tilde{\Delta}_3}_1}, \quad p_\text{SC} \in [0, 1]^K$, where $\Tilde{\Delta}_3$ is a vector composed of the SC values of all clients. The probability of selecting each client is proportional to its corresponding value in $p_\text{SC}$. 

\textbf{2.\,Complementary Selection (AI or CI):} After selecting a client based on SC, the next step ensures that the next selected client exhibits complementary data heterogeneity. To do so, the server computes the row-normalized matrix $\underline{\Tilde{\Delta}}$, where
    $\underline{\Tilde{\Delta}}_i^k = \frac{\Tilde{\Delta}_i^k}{\sum_{i=1}^3 \Tilde{\Delta}_i^k}\;, \quad \forall i \in \{ 1, 2, 3 \}, \, \forall k \in \mathcal{K}$. Using $\underline{\Tilde{\Delta}}$, the server selects the client whose normalized triplet is the least aligned (i.e. has the smallest dot product) with the normalized triplet of the already selected client. Formally, this is computed as:
    $k_c = \argmin_{k \in \mathcal{K} \setminus \{ k_p \}} \left\langle \underline{\Tilde{\Delta}}^{k_p}\,,\, \underline{\Tilde{\Delta}}^k \right\rangle$
    where $k_p$ denotes the already selected client and $\langle \cdot, \cdot \rangle$ represents the dot product. 

\textbf{3.\,Orthogonal Selection (CI or AI):} The next client $k_r$ is chosen to complement the heterogeneity profile of the data of the clients already selected. To achieve this, the server selects the client whose DHT aligns the most with the vector perpendicular to the DHTs of the two previously selected clients (which represent SC and either CI or AI). Formally, this is computed as: 
        $k_r = \argmax_{ k \in \mathcal{K} \setminus \{k_p, k_c \}} \left\langle \underline{\Tilde{\Delta}}^{k_p} \times \underline{\Tilde{\Delta}}^{k_c}, \underline{\Tilde{\Delta}}^{k} \right\rangle$
    where $(\cdot \times \cdot)$ is the cross product, ensuring that the selected client exhibits heterogeneity in the remaining dimension. 

This client selection approach leverages all three dimensions of the DHT by selecting clients with different types of data heterogeneity.
The server repeats the steps above iteratively until the desired number of clients has been selected, excluding clients already chosen in the current round. To enhance variability, the order in which dimensions (SC, CI, AI) are prioritized is rotated every three clients. 

As illustrated in the experimental section, \alg{FedDiverse}'s client selection can be applied in conjunction with any FL optimization approach.

\section{Experiments}
\label{sec:exp}
\subsection{Datasets}

\begin{table*}[ht]
    \caption{Worst group accuracies (mean and std) over three experiments of \alg{FedDiverse} and baselines in federation with 24 to 100 clients, with 9 clients selected every round, and \alg{FedAvgM} as the FL optimization algorithm. The best-performing method is \textbf{bold}, and the second best is \underline{underlined}. (*): 12 clients selected from 100. (**): Not scalable due to excessive computational cost. }
    \label{tab:selection}
    \centering
    \vspace{-0.5\baselineskip}
    \begin{adjustbox}{width=.8\linewidth}
        \centering
    
        \begin{tabular}{cccccccc}
        
            \toprule
            
            \multirow{2}{*}{\makecell{\textbf{Client Selection} \\ \textbf{algorithm}}}&\multicolumn{7}{c}{\textbf{Dataset}}\\
            &Spawrious\textsubscript{GSC}&Spawrious\textsubscript{GCI}&Spawrious\textsubscript{GAI}&WaterBirds\textsubscript{dist} &Spawrious\textsubscript{4} &CMNIST\textsubscript{GSC}  &Spawrious\textsubscript{GCI-100}*\\
            \midrule
            \alg{FedDiverse}     &\uum{88.01}{0.96}&\bum{89.91}{1.91}&\bum{87.28}{1.61}&\uum{54.10}{2.03}&\bum{86.06}{0.58}
            &\bum{94.01}{0.98}&\bum{91.22}{1.61}\\
            Uniform random       &\num{86.27}{1.12}&\num{87.59}{2.00}&\num{85.86}{2.56}&\num{42.42}{0.59}&\num{84.02}{0.63}&\num{92.00}{1.61}&\num{86.96}{1.28}\\
            Round robin          &\num{87.12}{0.87}&\num{87.64}{0.90}&\num{86.17}{2.65}&\num{41.23}{2.18}&\num{83.54}{1.83}&\uum{93.51}{0.49}&\num{85.54}{0.40}\\
            \alg{FedNova}        &\num{87.49}{0.73}&\num{88.52}{1.49}&\uum{87.22}{0.47}&\num{42.83}{0.71}&\num{84.65}{0.64}&\num{93.23}{0.34}&\num{87.33}{0.18}\\
            \alg{pow-d}         &\bum{89.12}{0.32}&\uum{89.01}{1.18}&\num{86.91}{1.52}&\bum{56.75}{2.49}&\num{83.54}{2.01}&\num{92.85}{0.47}&\uum{89.85}{1.00}\\
            \alg{FedPNS}         &\num{85.75}{1.34}&\num{85.02}{9.12}&\num{82.22}{6.94}&\num{48.75}{12.14}&\num{84.35}{1.45}&\num{91.49}{1.42}&N/A**\\
            \alg{HCSFed}         &\num{86.80}{0.86}&\num{87.17}{0.27}&\num{85.96}{2.70}&\num{41.66}{1.80}&\uum{85.59}{0.66}&\num{91.45}{1.11}&\num{85.49}{0.78}\\
            \bottomrule
    
        \end{tabular}
    \end{adjustbox}

\end{table*}

\begin{figure}
    \centering
    \begin{adjustbox}{width=.9\linewidth}
    \includegraphics{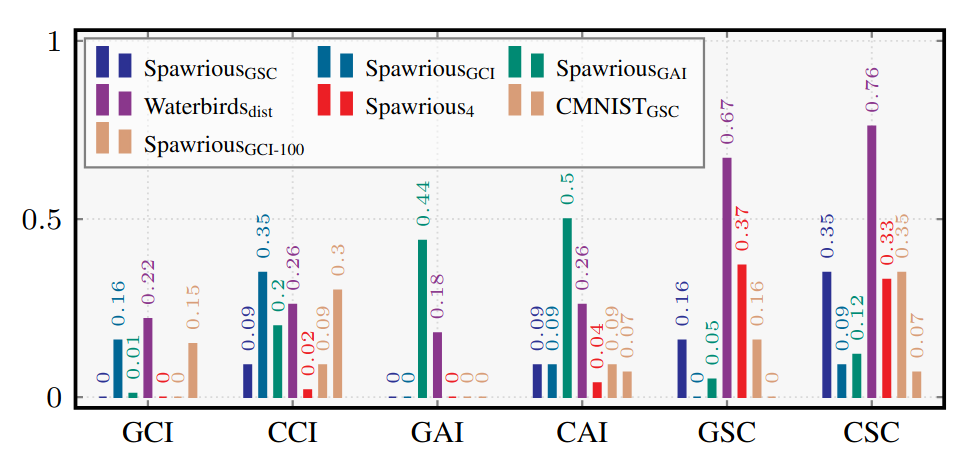}
\end{adjustbox}
\caption{
Global and client statistical data heterogeneity metrics of the proposed datasets. Note how each dataset has different values of class imbalance, attribute imbalance and spurious correlations both globally and locally.
}
\label{fig:datasplits}
\end{figure}

We perform the experimental evaluation 
on 7 different datasets that cover a wide variety of CI, AI and SC both globally and in the clients, as reflected in \cref{fig:datasplits} and described below.  

\textbf{1. WaterBirds:}
The WaterBirds dataset~\cite{wah2011caltech} is an image dataset of two classes and backgrounds with spurious correlation present between them.
We follow the original train/test split and distribute the training data over 30 clients as follows: 3 clients predominantly have CI; 2 clients have mostly AI; and the rest of the clients are impacted largely by the same SC as the global dataset.

\textbf{2. Spawrious:} The Spawrious dataset~\cite{lynch2023spawrious} consists of 4 dog breeds (target labels $y$) on 6 background (attributes $a$). There are 6,336 images for each $(y,a)$ pair, making it the largest vision dataset where the level of spurious correlation is adjustable~\cite{ye2024spurious}. We save $10\%$ of the data to create a balanced test set and use the remaining data to generate 5 federated datasets with various levels of statistical data heterogeneity. We identify and use the 2 hardest background groups (namely \textit{beach} and \textit{snow}) together with 2 (\textit{labrador} and \textit{dachshund}) or 4 dog breed classes.

While the WaterBirds dataset contains CI, AI and SC (see \cref{fig:datasplits}), we create \textbf{5 Spawrious} datasets with different data distributions to investigate the impact of CI, AI and SC individually: (1) 3 datasets where only one type of data heterogeneity is present globally, namely spurious correlation in Spawrious$_{GSC}$; class imbalance in Spawrious$_{GCI}$; and attribute imbalance in Spawrious$_{GAI}$; (2) Spawrious$_4$ with high levels of spurious correlation and 4 classes; and (3) Spawrious\textsubscript{GCI-100} with class imbalance and 100 clients. 

\textbf{3. CMNIST:}
The CMNIST dataset~\cite{arjovsky2019invariant} is generated based on the binarized MNIST dataset, with labels $y=0$ for digits less than five and $y=1$ otherwise. 
The attribute is given by the foreground color, $\mathcal A = \{\text{red}, \text{green}\}$. We use the same data distribution as for Spawrious$_{GSC}$ with 2 classes with high level of spurious correlations. 

\subsection{Experimental Setup}
We simulate a federated learning scenario with a total of 24 to 100 clients depending on the dataset\footnote{The federations with the Spawrious$_{GSC}$, Spawrious$_{GCI}$ and CMNIST$_{GSC}$ datasets have 24 clients; Spawrious$_{GAI}$ and Spawrious$_{4}$ have 25 clients; WaterBirds$_{dist}$ has 30 clients; and Spawrious$_{GCI-100}$ has 100 clients.} using the Flower~\cite{beutel2020flower} and PyTorch~\cite{paszke2019pytorch} frameworks.

The server and the clients trained a MobileNet v2~\cite{howard2017mobilenets} model, where batch normalization layers were replaced with group normalization layers and initial weights are pre-trained on Imagenet. We applied the \textit{categorical crossentropy} loss function with $0.001$ learning rate and a batch size of $28$. 
Unless otherwise noted, we used $T=200$ rounds of federated training with equally weighted clients. In experiments without client selection, all clients (24 to 100) participate in the federation in every round. In the cases where client selection is performed, the server selects $9$ clients to participate in the federation in each round, except for Spawrious\textsubscript{GCI-100} where $12$ clients are selected. 

We performed all experiments on the previously described datasets. We report \textit{worst-group accuracy}~\cite{SKHL2020} and its standard deviation, defined as 
 $   \min_{(y,a) \in \mathcal Y \times \mathcal A} \mathbb E[\mathbbm{1}\{y = f(x;\theta)\} \mid Y= y , A= a]$
over 3 runs using a balanced global test dataset. 

\subsection{Baselines}

We compare \alg{FedDiverse}'s client selection strategy with 6 baselines, described below. All the methods are implemented using server side momentum \alg{FedAvgM}~\cite{hsu2019measuring}. 

\textbf{1.\,Uniform random} selection, where clients are randomly selected according to a uniform distribution. 

\textbf{2.\,Round robin} selection, where the server keeps track of how many times $R_k$ a client $k$ has been selected such that the client cannot participate again while $\exists j\neq k, R_j<R_k$.
 
\textbf{3.\,\alg{FedNova}}~\cite{wang2020tackling}, a client weighting approach by means of importance weighting. The parameter aggregation is given by $\theta^{t+1}=\theta^t-\tau_{eff}\sum_{k}\frac{|D_k|}{|D|}\cdot \beta\nabla_k^{t+1}$, where $\beta$ is the same momentum as in \alg{FedAvgM} and $\tau_{eff}$ is the effective iteration step and it is computed from the client's steps. 

\textbf{4.\,\alg{pow-d}}~\cite{cho2022towards}, a loss-based selection method. First, the server $\mathcal{S}$ selects $k_{br}:\kappa<\kappa_{br}<K$ clients randomly to broadcast the model parameters $\theta^{t}$. All $k\in S_{\kappa_{br}}$ clients compute $\ell(\theta^t,D_k)$ and report it back to the server. Then, the server sorts the clients such that for $i,j\in \{1,\dots,K\}, i<j\rightarrow \ell(\theta^t,D_i)<\ell(\theta^t,D_j)$ and selects the first $\kappa$ clients to participate in the computation of $\theta^{t+1}$.

\textbf{5.\,\alg{FedPNS}}~\cite{wu2022node} identifies clients that negatively impact the aggregated gradient change by comparing a client's gradient change $\nabla^{t+1}_k$ with the overall gradient change excluding that client, $\nabla^{t+1}-\nabla^{t+1}_k$. If a client slows down the aggregated gradient, as indicated by 
$\langle \nabla^{t+1}, (\nabla^{t+1}-\nabla^{t+1}_k) \rangle$, the client is flagged. Flagged clients are less likely to be selected in subsequent rounds, while non-flagged clients and those not sampled in round $t$ are more likely to be selected.

\textbf{6.\,\alg{HCSFed}}~\cite{song2023fast} clusters the clients based on the compressed gradients after the first round of training. We use 3 clusters and randomly select clients from each cluster.

\subsection{Communication and Computation Overhead}

The baselines have varying levels of communication and computation overhead reported in Table \ref{tab:overhead}. \alg{FedNova} performs client weighting instead of selection, hence, all the clients participate in the federation in each round. While \alg{pow-d} performs client selection, the server needs $\ell(\theta^t,D_k)$ from all clients to determine which clients to select in each round. \alg{FedPNS} requires no additional work from the clients, but the server calculates the similarity between the client gradient updates in every round, which can result in significant overhead for complex models and large number of clients. \alg{HCSFed} addresses this issue by compressing the model gradients and organizing the clients into clusters after the first training round and minimizing the overhead for subsequent rounds.
Uniform random, Round robin and \alg{FedDiverse} are the only three client selection methods where \textbf{only the participating clients} perform computations and communicate with the server in a round. \alg{FedDiverse}'s additional communication overhead is limited to just 3 scalar values per client while the client-side computational overhead occurs only in a single training round. The only recurring overhead is the server-side selection, which involves sorting clients based on their DHT values.

\begin{table}[tp]
    \centering
    \caption{Communication and computation overhead for \alg{FedDiverse} and the baselines where $K=24..100, r=10^{-5}, |\theta|=2.23\cdot10^6, n_k=10^2..10^3$}
    \label{tab:overhead}
    \begin{adjustbox}{width=\linewidth}
    \begin{tabular}{ccccc}
    \toprule
       \multirow{2}{*}{Method}&\multirow{2}{*}{Frequency}&{Communication}&\multicolumn{2}{c}{Computation Overhead} \\
       & & Overhead & Client & Server ($\forall t)$ \\
    \midrule
        \alg{FedDiverse}&$t=1$&$3$&$\forall k\in\mathcal{K}:2 O(n_k|\theta|)$&$O(K)$\\
       Round Robin&0&0  & 0& $O(1)$\\
       \alg{FedNova}&$\forall t$&$3+\forall k\notin\mathcal{K}:\theta_k^t$&$O(1)+\forall k\notin\mathcal{K}:O(n_k|\theta|)$&$O(K)$\\
       \alg{pow-d}&$\forall t$&$1+\forall k\notin\mathcal{K}:\theta_k$&$\forall k\notin\mathcal{K}:O(n_k|\theta|)$&$O(K\log K)$\\
       \alg{FedPNS}&0&0&0&$O(K^2|\theta|^2)$\\
       \multirow{2}{*}{\alg{HCSFed}}&\multirow{2}{*}{$t=1$}&\multirow{2}{*}{$r\theta_k$}&\multirow{2}{*}{$r|\theta|^2$}&$t=1: O(K\cdot r|\theta|)$\\
       &&&&$t\neq 1: O(K)$\\
    \bottomrule
    \end{tabular}
    \end{adjustbox}
\end{table}





%

\begin{table*}[htp]
    \caption{Worst group accuracies (mean and std) over three experiments of \alg{FedDiverse} combined with four FL optimization methods on the proposed datasets vs the default random selection. The best-performing client selection method is highlighted in bold and the best-performing combination is underlined. Note how all the FL optimization algorithms improve their performance when doing client selection with \alg{FedDiverse} vs random selection across all datasets.}
    \label{tab:optimizers}
    \vspace{-0.5\baselineskip}
    \centering
    \begin{adjustbox}{width=\linewidth}
        \centering
        \setlength{\tabcolsep}{2pt}
        \begin{tabular}{ccccccccccccc}
        
            \toprule

            \multirow{2}{*}{\vspace{-.2em}\textbf{FL algorithm}} & \multicolumn{2}{c}{Spawrious\textsubscript{GSC}}& \multicolumn{2}{c}{Spawrious\textsubscript{GCI}}& \multicolumn{2}{c}{Spawrious\textsubscript{GAI}}& \multicolumn{2}{c}{WaterBirds\textsubscript{dist}}& \multicolumn{2}{c}{Spawrious\textsubscript{4}}& \multicolumn{2}{c}{CMNIST\textsubscript{GSC}} \\
            \cmidrule(r){2-3}\cmidrule(r){4-5}\cmidrule(r){6-7}\cmidrule(r){8-9}\cmidrule(r){10-11}\cmidrule(r){12-13}
            & Random & \alg{FedDiverse} & Random & \alg{FedDiverse} & Random & \alg{FedDiverse} & Random & \alg{FedDiverse} & Random & \alg{FedDiverse} & Random & \alg{FedDiverse} \\
            
            \midrule

            \alg{FedAvg}                  & \num{85.09}{1.00} & \bum{85.90}{1.62} & \num{85.65}{3.85} & \bum{89.43}{0.63}& \num{80.49}{0.52} & \bum{84.33}{1.51}& \num{31.72}{3.05} & \bum{46.47}{1.31}& \num{81.07}{1.29} & \bum{83.86}{1.20}& \num{87.58}{2.38} & \bum{91.01}{0.58}\\
            \alg{FedAvgM}                 & \num{86.27}{1.12} & \bum{88.01}{0.96} & \num{87.59}{2.00} & \bum{89.91}{1.91}& \num{85.86}{2.56} & \underline{\bum{87.28}{1.61}} & \num{42.42}{0.59} & \underline{\bum{54.10}{2.03}} & \underline{\bum{84.02}{0.63}} & \bum{86.06}{0.58} & \num{92.00}{1.61} & \underline{\bum{94.01}{0.98}} \\
            \alg{FedProx}                 & \num{84.43}{1.91} & \bum{86.33}{1.49} & \num{82.91}{4.89} & \bum{87.30}{3.01}& \num{81.39}{2.12} & \bum{83.81}{2.46}& \num{31.57}{2.87} & \bum{43.51}{0.70}& \num{80.44}{1.55} & \bum{83.64}{0.74}& \num{91.36}{0.95} & \bum{91.49}{1.86}\\
            \alg{FedAvgM} + \alg{FedProx} & \num{85.41}{1.67} & \underline{\bum{87.85}{1.26}} & \num{88.38}{1.42} & \underline{\bum{90.48}{1.61}} & \bum{85.65}{3.76} & \num{85.17}{1.79}& \num{44.29}{1.26} & \bum{53.84}{0.90}& \num{82.97}{0.69} & \underline{\bum{86.12}{0.97}}& \num{92.42}{0.71} & \bum{93.24}{0.38}\\

            \bottomrule
    
        \end{tabular}
    \end{adjustbox}

\end{table*}

\subsection{Results}

\cref{tab:selection} depicts the worst group accuracies for \alg{FedDiverse} and all the baselines on the 7 datasets. Note how client selection with \alg{FedDiverse} is the \emph{only method} that is the best or second best performing approach on \textbf{all datasets}. 

\subsubsection{Benchmarking FedDiverse with FL methods}
We evaluate \alg{FedDiverse}'s ability to improve the robustness of existing FL optimization algorithms when combined with them. We aim to (1) evaluate the ability of \alg{FedDiverse}'s client selection method to improve performance across a variety of datasets and FL optimization algorithms; and (2) shed light on which method yields the best performance. The algorithms benchmarked in this section are:

\textbf{1.\,\alg{FedAvg}}~\cite{mcmahan2017communication}, which  serves as the baseline FL method where in each round the global model is replaced by the average of the client models.

\textbf{2.\,\alg{FedAvgM}}~\cite{hsu2019measuring}, which includes server-level momentum. It is designed to improve non-IID convergence. The momentum parameter is set to $\beta=0.95$.

\textbf{3.\,\alg{FedProx}}~\cite{li2020federated}, where the client loss contains a proximal term derived from the difference between server and client weights to stabilize the convergence: $\ell_{prox}(f_k(x;\theta_k),y)=\ell(f_k(x;\theta_k),y)+\frac{\mu}{2}||\theta-\theta_k||_2$, where $\mu$ is a parameter set to $0.1$ in our experiments.

As \alg{FedAvgM} changes the server aggregation method, \alg{FedProx} the local loss function, and \alg{FedDiverse} the client selection policy, we can use any combination of the 3 methods to mitigate statistical data heterogeneity. As reflected in \cref{tab:optimizers}, \alg{FedDiverse} improves the performance over random selection when combined with every FL method and in all datasets. The combination of \alg{FedDiverse} with \alg{FedAvgM} yields very competitive performance and hence we opt for \alg{FedAvgM} as the FL optimization method to be used in all of the experiments.

\subsubsection{Ablation study}
In this section, we study the performance of \alg{FedDiverse} on the WaterBirds dataset and under different configurations, reflected in \cref{tab:steps}. 
We compare 3 scenarios: 

1. Our realistic setup, where the interaction matrix $\Tilde{N}^{k}$ and the data heterogeneity triplets $\Tilde{\Delta}^k$ are estimated;

2. An ideal --yet unrealistic-- scenario where the interaction matrix $N^{k}$ and therefore the triplets $\Delta^k$ are known to the server; and 

3. A method where the full interaction matrix $N^{k}$ is sent to the server instead of the triplets. In this case, the server first computes the client weights $\omega_k$ that minimize the variance of the matrix $\mathbf{S}=\sum_{k\in \mathcal K}\omega_{k}N^{k}$, \emph{i.e.}, 
$    \min \text{Var}(\mathbf{S})=\frac{1}{|Y||A|}\sum_{y\in Y}\sum_{a\in A}(s_{y,a}-\nu)^2$, where $\nu$ is the average number of samples per $(y,a)$ groups. We solve it as a convex optimization problem and use the $\omega_k$ weight as the probability to sample client $k$. Note that this method would raise privacy concerns.

Furthermore, we evaluate the impact of increasing the number of pre-training steps and compare \alg{FedDiverse} when combined with \alg{FedAvg} and \alg{FedAvgM}. 

As seen in the table, perfect knowledge of $N^{k}$ could yield an increase of up to $5.71$ and $3.74$ points in worst group accuracy with \alg{FedAvg} and \alg{FedAvgM}, respectively. Communicating the true (typically unknown) interaction matrix instead of the triplets could add up to $4.15$ and $4.78$ points to the worst-group accuracy with \alg{FedAvg} and \alg{FedAvgM}, respectively. Increasing the number of pre-training steps is only helpful with \alg{FedAvgM}, yet the performance gains are not significant.

\begin{table}[ht]    
    \caption{Ablation study of \alg{FedDiverse} with different configurations on the WaterBirds dataset.
    }
    \label{tab:steps}
    \vspace{-0.5\baselineskip}
    \centering
    \begin{adjustbox}{width=\linewidth}
        \centering
    
        \begin{tabular}{ccccc}
        
            \toprule

            \multirow{2}{*}{\textbf{Pre-training ($T_0$)}} & \multirow{2}{*}{\textbf{Interaction matrix}} & \multirow{2}{*}{\textbf{Message}} & \multicolumn{2}{c}{\textbf{Worst group accuracy (\%)}} \\
            &&&\alg{FedAvg}&\alg{FedAvgM}\\
            
            \midrule

            20 & predicted & DHT($\Tilde{\Delta}^k$) & \num{44.03}{0.32}&\num{55.04}{3.09}\\
            1 & predicted & DHT($\Tilde{\Delta}^k$) & \num{46.47}{1.31}&\num{54.10}{2.03}\\
            20 & known & DHT($\Delta^k$) & \num{48.08}{3.27}&\num{58.00}{0.77}\\
            1 & known & DHT($\Delta^k$) & \num{50.62}{3.00}&\num{58.88}{2.57}\\
            20 & known & $N^k$ & \num{49.74}{2.30}&\num{58.41}{2.04}\\
            1 & known & $N^k$ & \num{51.82}{3.79}&\num{57.84}{0.48}\\
            
            \bottomrule
    
        \end{tabular}
    \end{adjustbox}

\end{table}

\section{Conclusion}
\label{sec:concl}

In this work, we have introduced a novel framework for characterizing statistical data heterogeneity in FL, we have presented seven datasets to evaluate the performance of FL methods in the presence of different types of data heterogeneity, and we have proposed \alg{FedDiverse}, a novel and efficient client selection method that selects clients with diverse types of statistical data heterogeneity. In extensive experiments, we demonstrate the competitive performance of \alg{FedDiverse} on all datasets while requiring low communication and computation overhead.  

{
    \small
    \bibliographystyle{ieeetr}
    \bibliography{main}
}

\end{document}